\documentclass[letterpaper, 10 pt, conference]{ieeeconf}  

\IEEEoverridecommandlockouts                              

\usepackage{amsmath,amsfonts}
\usepackage{algorithm}
\usepackage{array}
\usepackage[caption=false,font=normalsize,labelfont=sf,textfont=sf]{subfig}
\usepackage{textcomp}
\usepackage{stfloats}
\usepackage{url}
\usepackage{verbatim}
\usepackage{cite}
\usepackage{graphicx} 
\usepackage{epsfig} 
\usepackage{tensor}
\usepackage{siunitx}

\usepackage{color}
\usepackage{times}
\usepackage{balance}
\usepackage{fancyhdr}
\usepackage{comment}
\usepackage{changepage}
\usepackage{bm}
\usepackage{calligra}
\usepackage{tabularx}
\usepackage{mathtools}
\usepackage{arydshln}
\usepackage{latexsym} 
\usepackage{float}
\usepackage{tabstackengine}
\usepackage{threeparttable}
\usepackage{subcaption}
\usepackage{xcolor}
\usepackage{booktabs}
\usepackage{algpseudocode}
\usepackage{multirow}
\usepackage{colortbl} 

\definecolor{lightgray}{rgb}{0.9, 0.9, 0.9} 

\usepackage{hyperref}
\hypersetup{
    colorlinks=true,
    linkcolor=subsectioncolor,
    filecolor=magenta,      
    urlcolor=cyan,
}

\DeclareMathOperator*{\argmax}{arg\,max}

\definecolor{subsectioncolor}{RGB}{255,0,0} 

\title{\LARGE \bf
MS-Mapping: Multi-session LiDAR Mapping with Wasserstein-based Keyframe Selection
}

\author{Xiangcheng Hu$^{1}$, Jin Wu$^{1}$, Jianhao Jiao$^{2}$,Wei Zhang$^{1}$ and Ping Tan$^{1}$
\thanks{$^{\dagger}$Corresponding Author}
\thanks{$^{1}$X. Hu, J. Wu, W. Zhang and P. Tan are with the Department of Electronic and Computer Engineering, Hong Kong University of Science and Technology, Hong Kong, China. E-mail: \tt\small{ pingtan@ust.hk}}%
\thanks{$^{2}$J. Jiao is with the Department of Computer Science, University College London, Gower Street, WC1E 6BT, London, UK. E-mail: \tt\small{ ucacjji@ucl.ac.uk}}%
}

\begin{document}

\maketitle
\thispagestyle{empty}
\pagestyle{empty}

\begin{abstract}

Large-scale multi-session LiDAR mapping is crucial for various applications but still faces significant challenges in data redundancy, memory consumption, and efficiency. 
This paper presents MS-Mapping, a novel multi-session LiDAR mapping system that incorporates an incremental mapping scheme to enable efficient map assembly in large-scale environments. 
To address the data redundancy and improve graph optimization efficiency caused by the vast amount of point cloud data, we introduce a real-time keyframe selection method based on the Wasserstein distance. Our approach formulates the LiDAR point cloud keyframe selection problem using a similarity method based on Gaussian mixture models (GMM) and addresses the real-time challenge by employing an incremental voxel update method.
To facilitate further research and development in the community, we make our code\footnote{https://github.com/JokerJohn/MS-Mapping} and datasets publicly available.
\end{abstract}

\section{INTRODUCTION}
\subsection{Motivation and Challenges}
Large-scale multi-session LiDAR mapping has become increasingly crucial for a wide range of applications, including surveying, autonomous driving, and multi-agent navigation \cite{zou2024lta}. However, current methods still struggle to address the challenges posed by the massive volume of LiDAR point cloud data, which leads to data redundancy, high memory consumption, and slow pose graph optimization.
These issues often result in severe performance degradation of the mapping system and hinder long-term scalability. Moreover, the presence of redundant data further exacerbates the computational complexity and memory overhead, making it challenging to perform large-scale mapping on resource-constrained platforms.
Researchers have proposed direct solutions for pose graph optimization, such as pose graph pruning or sparsification\cite{mal2016Nonli}. However, these methods are often complex and suffer from accuracy loss while struggling to run in real-time. An alternative approach is to judiciously select keyframes, which both reduces data redundancy and  pose graph size.
Some optimization-based methods define certain metrics to select the optimal combination of keyframes from a set. However, these approaches generally lack real-time performance. Information theory-based methods attempt to determine keyframes based on changes in feature point information content, but they require calculating the covariance of points in the neighborhood, which is computationally expensive due to the large volume of point cloud data.
Consequently, current mainstream LiDAR keyframe selection methods typically set distance and rotation angle thresholds, selecting keyframes based on motion distance and angle. While fast, this spatial filtering approach ignores temporal scene changes and may result in information loss.

Drawing inspiration from information theory, we model the LiDAR keyframe selection problem as a similarity measure between two Gaussian mixture distributions (GMM). We observe the map distribution before and after adding a point cloud frame to select keyframes. The Wasserstein distance-based method helps us effectively capture both global and local differences between map distributions. To improve computational efficiency, we divide the map into voxels and use an incremental voxel update method to update Gaussian parameters in the GMM map.
This paper emphasizes the following aspects:
\begin{itemize}
\item We propose MS-Mapping, a novel incremental multi-session LiDAR mapping algorithm that employs an incremental mapping scheme, enabling efficient map assembly in large-scale environments.
\item We formulate the LiDAR keyframe selection problem using GMM similarity for the first time, combined with an incremental voxel update method to achieve real-time updates of Gaussian parameters, thereby capturing the impact of keyframes on subtle changes in the map.
\item We introduce a real-time keyframe selection method based on the Wasserstein distance to reduce data redundancy and pose graph complexity.
\end{itemize}

\section{Problem Formulation}
In this section, we assume that the reader is familiar with the basic theories of state estimation, such as maximum a posteriori (MAP) estimation and factor graphs\cite{dellaert2012factor}.

Let $\mathcal{W}$ denote the common world frame. The transformation from the body frame (IMU frame) $\mathcal{B}$ to the world frame at time $t_i$ is represented as:
$\mathbf{T}_{b_i}^w = \begin{bmatrix}
\mathbf{R}_{b_i}^w & \mathbf{t}_{b_i}^w \\
\mathbf{0}^T & 1
\end{bmatrix} \in SE(3),$
where $\mathbf{R}_{b_i}^w \in SO(3)$ is the rotation matrix and $\mathbf{t}_{b_i}^w \in \mathbb{R}^3$ is the translation vector.
Given two data sequences $\mathcal{S}_1$ and $\mathcal{S}_2$, we define their trajectories as $\mathbf{X}_1 = \{\mathbf{T}_1^{w_1}, \mathbf{T}_2^{w_1}, \dots, \mathbf{T}_n^{w_1}\}$ and $\mathbf{X}_2 = \{\mathbf{T}_1^{w_2}, \mathbf{T}_2^{w_2}, \dots, \mathbf{T}_m^{w_2}\}$, respectively. The corresponding point cloud frames are denoted as $\mathbf{P}_1 = \{^1P_1, ^1P_2, \dots, ^1P_n\}$ and $\mathbf{P}_2 = \{^2P_1, ^2P_2, \dots, ^2P_m\}$, where the left superscript indicates the sequence and the subscript represents the frame index. The merged pose graph is represented as $\mathbf{G} = (\mathbf{V}, \mathbf{E})$, where $\mathbf{V} = \mathbf{X}_1 \cup \mathbf{X}_2$ is the set of pose nodes and $\mathbf{E}$ is the set of edges, including odometry edges and loop closure edges.

In multi-session mapping, we aim to incrementally expand the pose graph and construct a globally consistent map by optimizing the poses of $\mathcal{S}_2$ based on the existing pose graph of $\mathcal{S}_1$. This is achieved by incorporating LiDAR odometry and loop closure constraints between the two data sequences. We prioritize relative pose constraints\cite{hu2024paloc} from LiDAR odometry and emphasize loop closure constraints to ensure global consistency and accuracy for multi-session data captured by sensors with similar noise levels over short periods.
Given an initial guess $\mathbf{T}_{init}$ between $\mathcal{S}_1$ and $\mathcal{S}_2$, the problem can be formulated as estimating the optimal pose trajectory $\mathbf{X}_2$ that minimizes the overall error of the merged pose graph $\mathbf{G}$:
\begin{equation}
\arg\min_{\mathbf{X}_2} \sum_{i \in \mathcal{E}_{odo}^{\mathbf{G}}} \rho(\mathbf{e}_{odo_i}^{\mathbf{G}}) + \sum_{j \in \mathcal{E}_{lc}^{\mathbf{G}}} \rho(\mathbf{e}_{lc_j}^{\mathbf{G}}) + \sum_{k \in \mathcal{E}_{prior}^{\mathbf{G}}} \rho(\mathbf{e}_{prior_k}^{\mathbf{G}}),
\end{equation}
where $\mathcal{E}_{odo}^{\mathbf{G}}$, $\mathcal{E}_{lc}^{\mathbf{G}}$, and $\mathcal{E}_{prior}^{\mathbf{G}}$ denote the sets of odometry edges, loop closure edges, and prior factor in the merged pose graph $\mathbf{G}$, $\mathbf{e}_{odo_i}^{\mathbf{G}}$, $\mathbf{e}_{lc_j}^{\mathbf{G}}$, and $\mathbf{e}_{prior_k}^{\mathbf{G}}$ are the corresponding edge errors, and $\rho(\cdot)$ is a robust kernel function.





\section{Wasserstein-based Keyframe Selection}\label{sec:ws_keyframe}


\subsection{Keyframe Selection Formulation}\label{sec:sub_keyframe}

We formulate the keyframe selection problem as a map distribution similarity task. Intuitively, if a point cloud frame $P$, when added to the map $M$, can bring significant map changes, it will be considered as a keyframe. Let $M_1$ and $M_2$ denote the map distributions before and after the update of the new frame, respectively. The keyframe selection problem can be transformed into measuring the dissimilarity between two distributions, as shown in (\ref{eq:keyframe_formulation}).
\begin{equation}\label{eq:keyframe_formulation}
\text{KF} = \argmax_{P} \textbf{Distance}(M_1, M_2)
\end{equation}
The map distribution can be modeled as a Gaussian Mixture Model:
$M = \sum_{i=1}^{K} w_i \cdot \mathcal{N}(\mu_i, \Sigma_i)$,
\noindent where $K$ is the number of components, $w_i$ is the weight of the $i$-th component, and $\mathcal{N}(\mu_i, \Sigma_i)$ is the Gaussian distribution with mean $\mu_i$ and covariance $\Sigma_i$.
When a new point cloud frame $P$ is available, it is first transformed into the map coordinate system using the estimated pose $T \in SE(3)$, $P_M = T \cdot P$.
To ensure the reliability of the keyframe selection process, we make two assumptions. The pose $T$ and the map $M$ are accurately estimated, and the map $M$ completely covers the range of the transformed frame $P_M$.
The first assumption ensures a consistent scale between $M_1$ and $M_2$, while the second assumption avoids the interference of newly explored areas on the map information.

\begin{algorithm}[!ht]
\caption{Wasserstein-based Keyframe Selection}
\label{alg:keyframe_selection}
\begin{algorithmic}[1]
\Require Point cloud frame $P$, pose $T$, Wasserstein distance threshold $\tau$, voxel size $l$
\Ensure Keyframe decision $b \in {0, 1}$ for frame $P$
\State Initialize voxel map $M_1$
\For{each voxel $V_i$ in $M_1$}
\State Compute and store $\mu_i$, $\Sigma_i$, and points count $n_i$ using (\ref{eq:gaussian_u}), (\ref{eq:gaussian_cov}), and (\ref{eq:hash})
\EndFor
\While{frame $P$ is available}
\State Transform $P$ into the map coordinate $P_M$
\State Initialize voxel map $M_2$
\For{each point $p_M \in P_M$}
\State Incremental update $M_2$ using (\ref{eq:hash}), (\ref{eq:gaussin_new_u}), and (\ref{eq:gaussin_new_cov})
\EndFor
\State $D_W \gets \textsc{ComputeWassersteinDistance}(M_1, M_2)$
\If{$D_W > \tau$}
\State $b \gets 1$ \Comment{Mark frame $P$ as a keyframe}
\Else
\State $b \gets 0$ \Comment{Mark frame $P$ as a non-keyframe}
\EndIf
\State Remove outside voxels and update map $M_1$ and $M_2$
\EndWhile
\end{algorithmic}
\end{algorithm}

\begin{figure}
    \centering
    \includegraphics[width=0.48\textwidth]{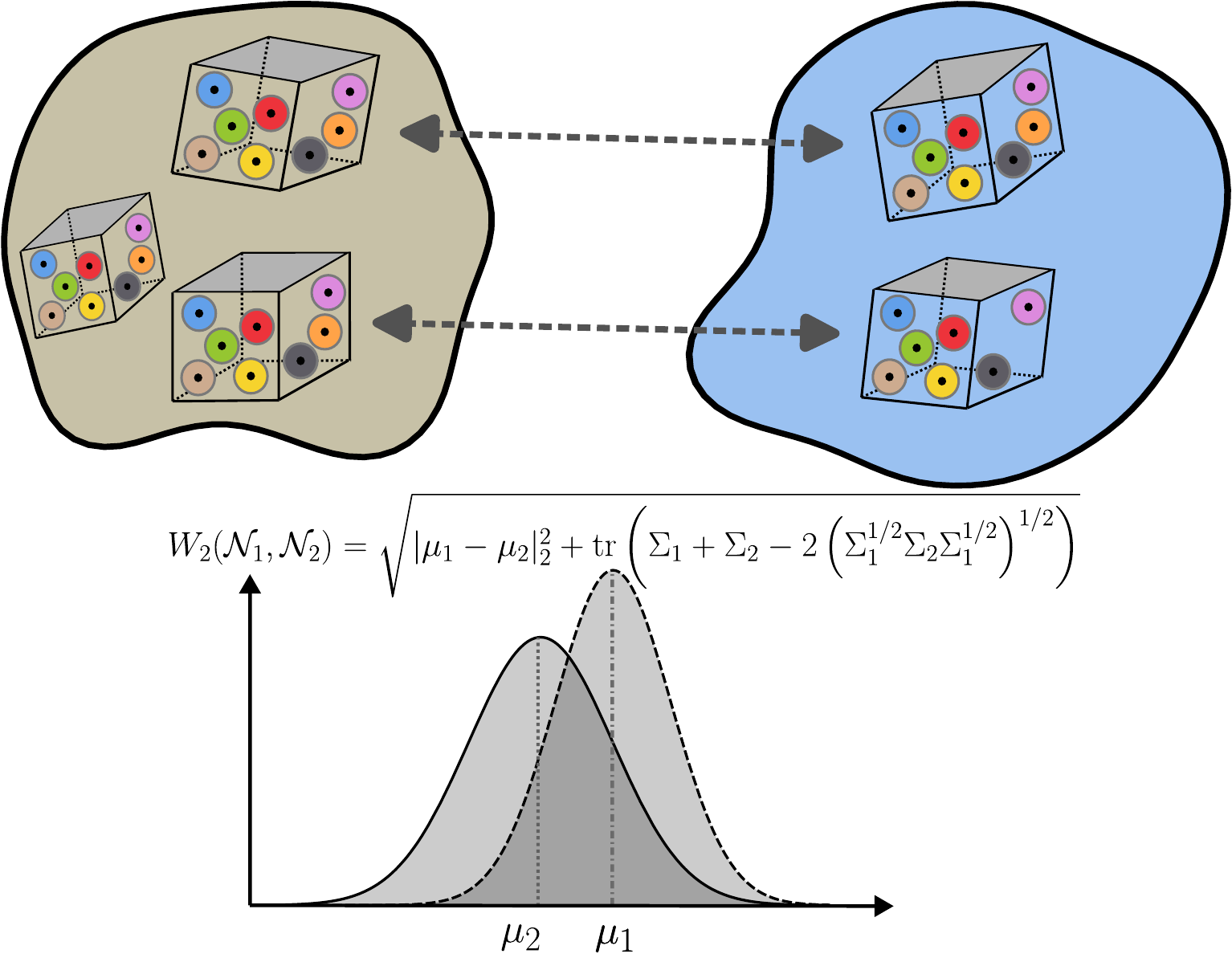}
\caption{Measuring differences between two point cloud maps using Wasserstein distance. The Wasserstein distance includes residuals of both the mean and variance, capturing both global and local differences between distributions. After voxelizing the point cloud maps, the number of points in each voxel represents its "mass," allowing the calculation of Wasserstein distance between voxel pairs. The overall GMM map difference is the average Wasserstein distance of all voxels. Thus, by comparing the average Wasserstein distance before and after adding a point cloud frame to the map, it can be determined if the frame is a keyframe.}
    \label{fig:wasserstein}
\end{figure}

\subsection{Wasserstein Distance for Map Dissimilarity}\label{sec:sub_wd}

To measure the dissimilarity between $M_1$ and $M_2$, we propose to use the Wasserstein distance, which is a theoretically optimal transport distance between two probability distributions. The L2 Wasserstein distance between two Gaussian distributions $\mathcal{N}_1(\mu_1, \Sigma_1)$ and $\mathcal{N}_2(\mu_2, \Sigma_2)$ is defined as:
\begin{equation}\label{eq:wasserstein_distance}
\scriptsize
W_2(\mathcal{N}_1, \mathcal{N}_2) = \sqrt{|\mu_1 - \mu_2|_2^2 + 
\text{tr}\left(\Sigma_1 + \Sigma_2 - 2\left(\Sigma_1^{1/2} \Sigma_2 \Sigma_1^{1/2}\right)^{1/2}\right)}
\end{equation}

The first term, $|\mu_1 - \mu_2|_2^2$, represents the squared Euclidean distance between the means, capturing the spatial distance between the distribution centers and indicating the overall shift in the map. The second term involves the trace of the covariance matrices and measures the dissimilarity in the shape and spread of the distributions, quantifying the difference in the local geometric structure of the map. By combining both the spatial shift and local geometric variation, the Wasserstein distance provides a comprehensive measure of map dissimilarity. Figure~\ref{fig:wasserstein} illustrates this process.

\subsection{Voxel Gaussian and Incremental Update}\label{sec:sub_voxel}

To manage the computational complexity of a large-scale point cloud map, we divide the local map into voxels of fixed size (e.g., \SI{4.0}{m}) and employ incremental updates for voxels affected by the new point cloud frame $P$. Each voxel is assumed to follow a Gaussian distribution, and the voxels are assumed to be independent of each other. Thus, the map distribution can be approximated as a GMM with each voxel being Gaussian distribution.

Let $V_i$ denote the $i$-th voxel, and $\mathcal{N}_i(\mu_i, \Sigma_i)$ be its corresponding Gaussian distribution. The mean $\mu_i$ and covariance $\Sigma_i$ can be computed as:
\begin{equation}\label{eq:gaussian_u}
\mu_i = \frac{1}{|V_i|} \sum_{x \in V_i} x,
\end{equation}
\begin{equation}\label{eq:gaussian_cov}
\Sigma_i = \frac{1}{|V_i| - 1} \sum_{x \in V_i} (x - \mu_i)(x - \mu_i)^T,
\end{equation}
\noindent where $|V_i|$ denotes the number of points in the $i$-th voxel.

\begin{figure}
    \centering
    \includegraphics[width=0.48\textwidth]{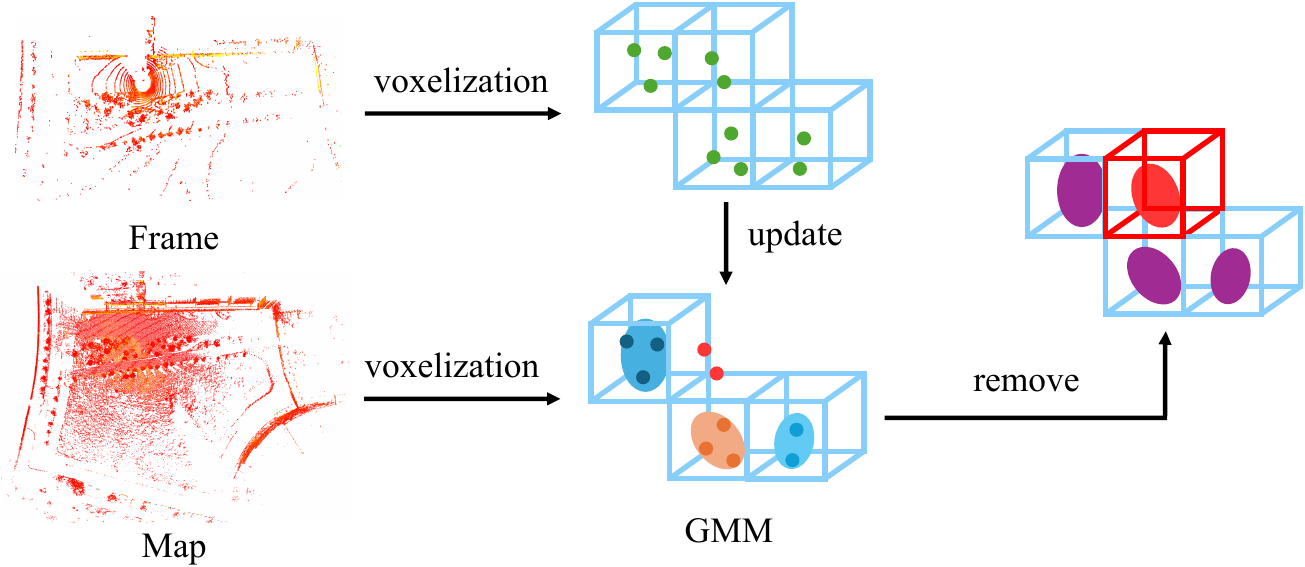}
\caption{Incremental update and maintenance of the GMM map. After initializing the GMM map, each new point cloud frame is transformed into the map coordinate. The Gaussian parameters of the voxels in the GMM are then updated point by point according to their indices. Voxels outside the radius range are removed. This process reflects the influence of the new frame on the overall map distribution.}
    \label{fig:voxel}
    \vspace{-0.5cm}
\end{figure}

When a new point cloud frame $P$ is added to the map, for each point $p \in P_M$, we compute its corresponding voxel index $(i, j, k)$ using:
\begin{equation}\label{eq:hash}
i = \left\lfloor \frac{p_x}{l} \right\rfloor, \quad j = \left\lfloor \frac{p_y}{l} \right\rfloor, \quad k = \left\lfloor \frac{p_z}{l} \right\rfloor,
\end{equation}
\noindent where $l$ is the voxel size. The Gaussian distribution of the corresponding voxel can be updated incrementally:
\begin{equation}\label{eq:gaussin_new_u}
\mu_i' = \frac{|V_i| \mu_i + \sum_{p \in V_i'} p}{|V_i| + |V_i'|},
\end{equation}
\begin{equation}\label{eq:gaussin_new_cov}
\Sigma_i' = \frac{|V_i| \Sigma_i + \sum_{p \in V_i'} (p - \mu_i')(p - \mu_i')^T}{|V_i| + |V_i'|}.
\end{equation}
\noindent where $V_i'$ denotes the set of new points falling into the $i$-th voxel. To efficiently retrieve the voxels, we use a hash table with the voxel index $(i, j, k)$ as the key and the voxel information as the value. This allows for fast updates and queries of the voxel-based map distribution. Figure~\ref{fig:voxel} illustrates the incremental GMM map update process.

\subsection{Real-time Keyframe Selection}
Algorithm \ref{alg:keyframe_selection} presents the Wasserstein distance-based keyframe selection process. 
The algorithm takes a point cloud frame $P$, its estimated pose $T$, a predefined Wasserstein distance threshold $\tau$, and the voxel size $l$ as inputs. It first initializes a voxel map $M_1$ and computes the Gaussian parameters and point count for each voxel. When a new frame arrives, it is transformed into the map coordinate system and used to update the voxel map incrementally. The average Wasserstein distance between the corresponding voxels in the previous and updated maps is computed using Equation \ref{eq:wasserstein_distance}. If the average distance exceeds the threshold $\tau$, the frame is marked as a keyframe. The algorithm continues to process incoming point cloud frames in real-time, effectively identifying informative keyframes for large-scale LiDAR mapping.

\section{CONCLUSIONS}

We presented MS-Mapping, a novel multi-session LiDAR mapping system that addresses the challenges of data redundancy through a real-time keyframe selection method. However, this keyframe selection method inevitably introduces additional computation time. Future work could focus on further improving performance.

\addtolength{\textheight}{-12cm}   





\bibliographystyle{IEEEtran}
\bibliography{ref}

\end{document}